\documentclass[11pt,a4paper]{article}
\usepackage[hyperref]{emnlp2020}
\usepackage{times}
\usepackage{latexsym}

\usepackage{microtype}

\aclfinalcopy 
\def\aclpaperid{2867} 

\usepackage{graphicx}
\usepackage{makecell}
\usepackage{tabularx}
\usepackage{multirow}
\usepackage{subfig}
\usepackage{colortbl}
\usepackage{amsmath}

\DeclareMathOperator*{\argmin}{argmin}
\usepackage[symbol]{footmisc}
\usepackage{hyperref}
\usepackage{booktabs}

\definecolor{earthquake}{HTML}{ffcc99}
\definecolor{flood}{HTML}{94c9ff}
\definecolor{typhoon}{HTML}{99ff99}
\definecolor{attack}{HTML}{f0d16e}
\definecolor{wildfire}{HTML}{ff6666}

\usepackage{amsmath}

\title{Event-Related Bias Removal for Real-time Disaster Events}

\author{Evangelia Spiliopoulou$^*$ \quad
Salvador Medina Maza$^*$  \quad
Eduard Hovy \quad
Alexander Hauptmann\\
Language Technologies Institute\\
Carnegie Mellon University\\
\texttt{\{espiliop,salvadom,ehovy,alex\}@cs.cmu.edu}}

\date{}

\begin{document}
\maketitle
\let\thefootnote\relax\footnote{\textsuperscript{*} Equal contribution.}

\begin{abstract}
Social media has become an important tool to share information about crisis events such as natural disasters and mass attacks. Detecting actionable posts that contain useful information requires rapid analysis of huge volume of data in real-time. This poses a complex problem due to the large amount of posts that do not contain any actionable information. Furthermore, the classification of information in real-time systems requires training on out-of-domain data, as we do not have any data from a new emerging crisis. Prior work focuses on models pre-trained on similar event types. However, those models capture unnecessary event-specific biases, like the location of the event, which affect the generalizability and performance of the classifiers on new unseen data from an emerging new event. In our work, we train an adversarial neural model to remove latent event-specific biases and improve the performance on tweet importance classification. 

\end{abstract}

\section{Introduction}
Effective management of crisis situations like natural disasters (e.g. earthquakes, floods) or attacks (e.g. bombings, shootings) is an extremely sensitive and complex phenomenon that requires efficient coordination of people from multiple disciplines along with proper allocation of time and resources \cite{tapia2011exploring,maitland2009information}. Given that we live in the era of information and social media, filtering important nuggets of information from real-time data and using them into decision-making constitutes a crucial research direction \cite{tapia2011seeking}.  

Critical information from social media is found only in small amounts. Hence it is difficult to extract and analyze the data stream, since it is impossible to manually process the amount of information shared in social media in real-time. Therefore, it is important to detect data that contain useful information for decision-making and automatically extract it \cite{sutton2008backchannels,palen2010vision}. Even though sentence classification is a well-studied NLP problem, common approaches do not bring the expected results \cite{reuter2018iscram}.

The main reason why common approaches fail is the lack of in-domain data \cite{mccreadie2019trec,hiltz2014use}. Most emerging crisis are unexpected and data analysis must be done real-time, within a small time-frame \cite{plotnick2015red}. Even if we might have high quality annotated data from previous similar crisis situations, we will not have data from the emerging event that we want to classify. For example, let us assume an earthquake in Seattle happens right now. Although we may have annotated data from a previous earthquake in Los Angeles, most of the parameters would be entirely different (e.g. location names, damages, times, etc) since the cities and populations differ. Furthermore, because some of those parameters might indeed play an important role in the classification of a tweet from the specific event (e.g. location, if Monroe is the epicenter of the Seattle earthquake), a traditional model would learn them as important features. This creates a highly-biased model that does not generalize on future events, since we cannot fine-tune properly on-the-fly. On the other hand, some other features are actually important in the general setting (e.g. severity of the earthquake, casualties etc.). The problem we tackle in this work is how to construct an event-based zero-shot learning model that can learn unbiased representations, instead of relying on a highly-biased set of features from seen data.

In this paper we explore a technique that helps a neural model to distinguish and discard information that is related only to specific events, resulting in a more generalizable model with improved performance on unseen events without any fine-tuning. Since the main task is to classify the importance of the information contained in a tweet (criticality), we use an adversarial classifier that intends to learn which specific event the tweet refers to, hence remove the event specific bias through a reversal gradient. Our experiments represent a real-life crisis management scenario, where the model is evaluated on a new incoming event through a leave-one-out experimental setup, and show substantial improvement over baseline classification methods. Finally, we share our code for reproducibility and ease of use$^1$\footnote{$^1$ \href{https://salmedina.github.io/EventBiasRemoval/}{https://salmedina.github.io/EventBiasRemoval/}}.

\section{Related Work}

Recent work on crisis informatics focuses on developing NLP solutions to classify and extract information from Twitter streams and other social media data related to an emergency event (e.g. attacks, natural disasters). As discussed by \citet{tapia2011seeking}, there are several problems under the umbrella of crisis informatics, such as determining if a snippet of text is related to a specific event, if it is reliable and trustworthy, the type of information it contains, whether the information is actionable, etc. Most previous work focuses on the relevance problem: given a set of tweets or other source of information and a specific event, classify which data refer to that event. \citet{caragea2016identifying} uses a CNN model to classify tweets related to flood events, while \citet{kruspe2019few} uses a few-shot learning model based on a CNN. \citet{nguyen2016applications} also uses a CNN model to classify related tweets and the type of information contained (e.g. infrastructure damage, affected individuals etc) from the Nepal 2015 earthquake. \citet{neubig2011safety} introduces a real-time system for the Japan 2011 earthquake that classifies the relatedness of the posts and extracts surface information like named entities. Other approaches include BiLSTM models for tweet classification \cite{matweets}, event detection based on Twitter streams \cite{sakaki2010earthquake}, adversarial data augmentation for image classification \cite{pouyanfar2019unconstrained} and domain-adaptation across different events using an adversarial network.

It is particularly important to first responders the identification of actionable information from a stream of messages as the one provided by Twitter. \citet{munro2011subword} proposes a system based on a set of features (location, time, n-grams) to label text messages as actionable/ non-actionable. Most recently, the TREC-IS challenge by \citet{mccreadie2019trec} proposes a labeling scheme where the actionability of a tweet is replaced by the information type and the criticality score. Higher criticality indicates a post contains more relevant information that could be useful for public safety officers during an emergency. Although \citet{miyazaki2019label} shows a great improvement on information type extraction by using Bi-LSTM attention on BERT embeddings, identifying critical and actionable information is a much harder task \cite{mccreadie2019trec}.    

Processing information without the context of a crisis event is a bottleneck for big data crisis analytics, as discussed by \citet{qadir2016crisis}. The lack of context makes the classification of messages very difficult, since the models are prone to event-specific biases. Due to the fact that we deal with real-time data, a domain-adaptation approach cannot use fine-tuning in a zero-shot scenario, which results in highly-biased models. Most recent work on bias removal \cite{elazar2018adversarial} focuses on using adversarial learning to remove demographic bias from representations. Examples include adversarial generative networks that create fair representations \cite{madras2018learning}, metrics to quantify unintended biases \cite{borkan2019nuanced} and applications that show substantial improvements on traditional NLP tasks like NLI \cite{lu2018gender}, Coreference Resolution \cite{belinkov2019adversarial} and text classification \cite{zhang2018mitigating} by using unbiased representations. Our approach is inspired by the work of \citet{elazar2018adversarial} on bias removal through an adversarial attack. The authors use an adversarial setting to remove demographic information from text and construct cleaner representations. In our case, the adversarial classifier attempts to predict the event to which the tweet belongs. Another difference with our work is the imbalanced data used for training the classifier of the main task. Other related work includes domain adaptation based on a gradient-reversal layer \cite{ganin2016domain}, text classification based on adversarial multi-task learning \cite{liu2017adversarial}, and multi-adversarial domain adaptation across multi-modal data \cite{pei2018multi}.

\section{Approach}

\begin{figure*}[ht]
    \centering
    \includegraphics[width=0.7\textwidth]{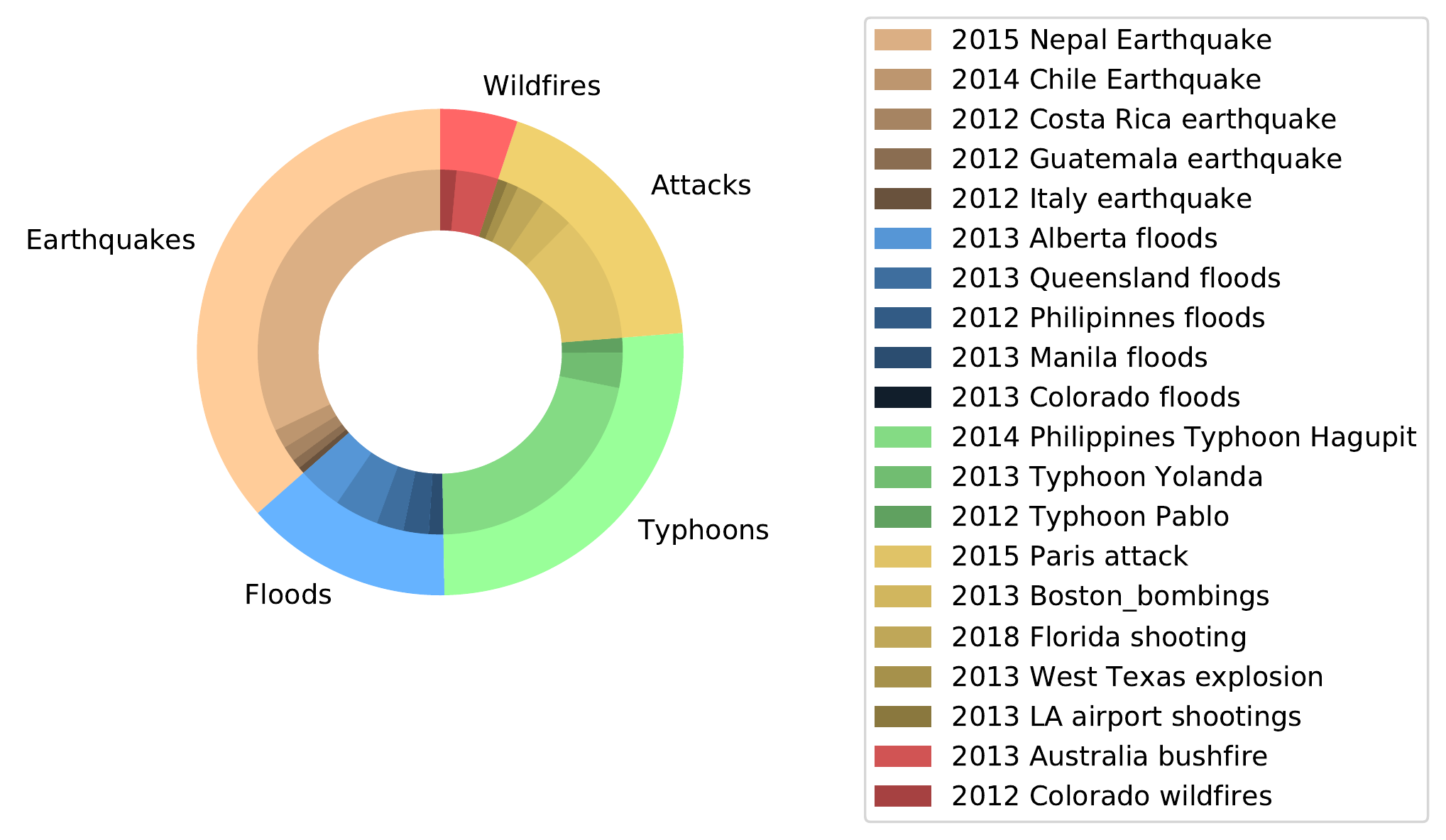}
    \caption{Crisis NLP Dataset Distribution. Outer circle: Color defines each of the event categories. Inner circle: The shade of colors describe the different events within a category.}
    \label{fig:dataset_pie}
\end{figure*}

In this work we used data from the TREC 2018 Incident Streams challenge$^2$\footnote{$^2$ \href{http://dcs.gla.ac.uk/~richardm/TREC\_IS/2020/oldindex.html}{http://dcs.gla.ac.uk/~richardm/TREC\_IS/2020/oldindex.html}}, which contains labels on criticality and information types \cite{mccreadie2019trec}. They define criticality as a score to identify posts that need to be shown to an officer immediately as an alert. The raw data and information about the specific event each tweet belongs to is extracted from the Crisis NLP \cite{imran2016lrec} dataset, which contains tweets in English from disaster events that occurred during 2012-2018. The crisis events in our dataset can be split into five main groups: earthquakes, floods, typhoons, wildfires and attacks. In Figure \ref{fig:dataset_pie}, we show that the data mainly consists of multiple earthquake, flood, and typhoon events, only two wildfire events, and five diverse attacks originated by humans.

\subsection{Data Description}
\begin{table}[h]
\caption{Examples of Critical and Non-Critical Tweets}
\label{tab:tweet_examples}
\resizebox{\columnwidth}{!}{
    \begin{tabular}{lll}
    \toprule
    \textbf{Label} & \textbf{Event} & \multicolumn{1}{l}{\textbf{Tweet}} \\ \midrule
    \rowcolor[HTML]{EFEFEF} 
    non-critical & 2014 Philippines Typhoon & Good morning! keep safe everyone! \\
    critical & 2013 Colorado Floods & \begin{tabular}[c]{@{}l@{}}RT: Seek higher ground immediately\\ wall of water coming down Boulder Canyon\\ move away from Boulder Creek\end{tabular} \\
    \rowcolor[HTML]{EFEFEF} 
    non-critical & 2013 Boston Bombings & \begin{tabular}[c]{@{}l@{}}I am honestly sick who could be so\\ disgusting to do this to someone we will get\\ answers and find you \#prayforboston\end{tabular} \\
    critical & 2015 Nepal Earthquake & \begin{tabular}[c]{@{}l@{}}RT: News at epicenter of Nepal tragedy\\ local church mission offers help!\end{tabular} \\ 
    \bottomrule
    \end{tabular}
}
\end{table}

\begin{figure}[htp] 
    \centering
    \subfloat[Original labels]{%
        \includegraphics[width=0.5\columnwidth]{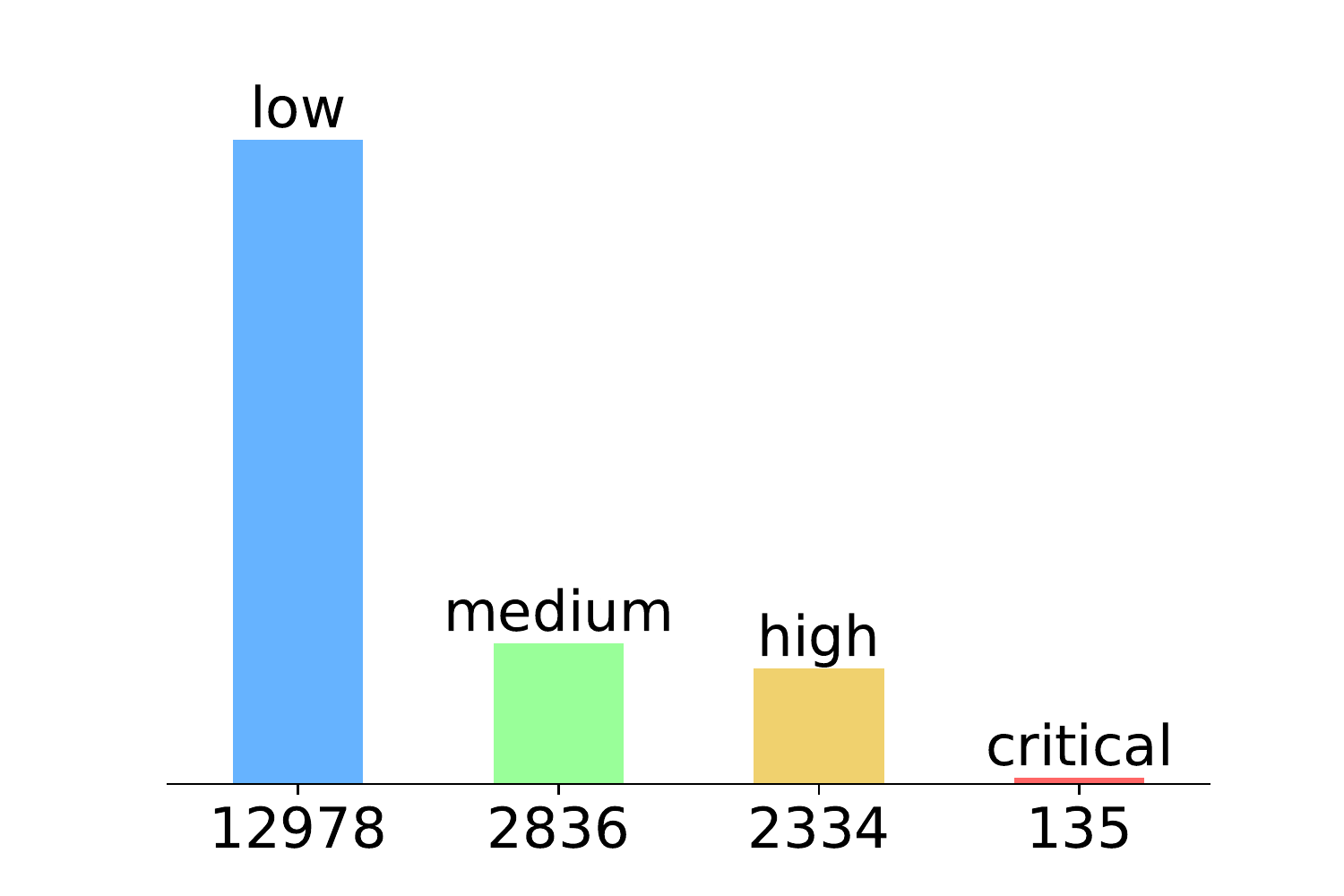}%
        \label{fig:original_labels_barplot}%
        }%
    \hfill%
    \subfloat[New labels]{%
        \includegraphics[width=0.5\columnwidth]{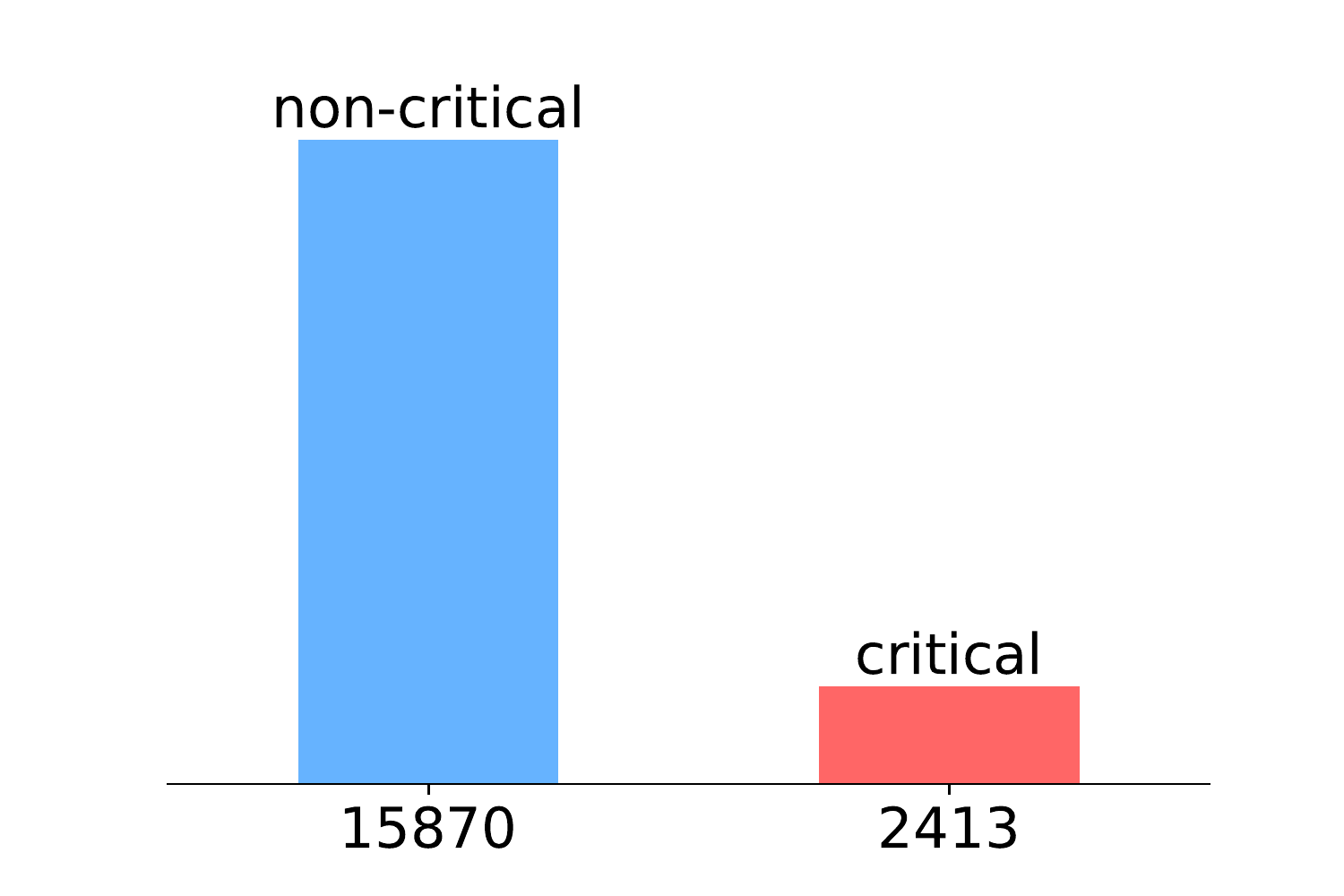}%
        \label{fig:new_labels_barplot}%
        }%
    \caption{Dataset label distribution. (a) Label distribution of original dataset, (b) Distribution of the labels after grouping \{low, medium\} as non-critical and \{high, critical\} as critical}
\end{figure}

In our experiments we used a labeled subset of the data formed by 18,283 tweets which are labeled into four categories according to their level of importance for the authorities: low, medium, high, and critical. The distribution of the labels is highly skewed towards the low and medium labels as shown in Figure \ref{fig:original_labels_barplot}. These types of tweets do not provide important information for decision-making during a disaster event. Since we are aiming to sieve the actionable tweets, we grouped together the low and medium labels as \textit{non-critical}, and the high and critical as \textit{critical}. The new distribution of the data after relabeling is shown in Figure \ref{fig:new_labels_barplot}. As we see on the examples shown in Table \ref{tab:tweet_examples}, the latter have actionable information for the authorities, first responders, and population on distress.

\subsection{Data Pre-processing}

Our target dataset comes from Twitter. Therefore, we performed a series of pre-processing steps for data-cleaning. First, we removed links, hashtags and mentions, since most of them are event specific. We also removed non-English words to reduce the noise. Next, we removed all non-English characters and emojis. Finally, we observed that many times white spaces were omitted between words, which resulted in multiple words being clustered as a single token. To solve that, we stripped the text from punctuation marks and, subsequently, used a heuristic for word segmentation, where we split the token into the least number of possible English words via greedy search.        


\subsection{Models}

\begin{figure}[ht]
    \centering
    \includegraphics[width=\columnwidth]{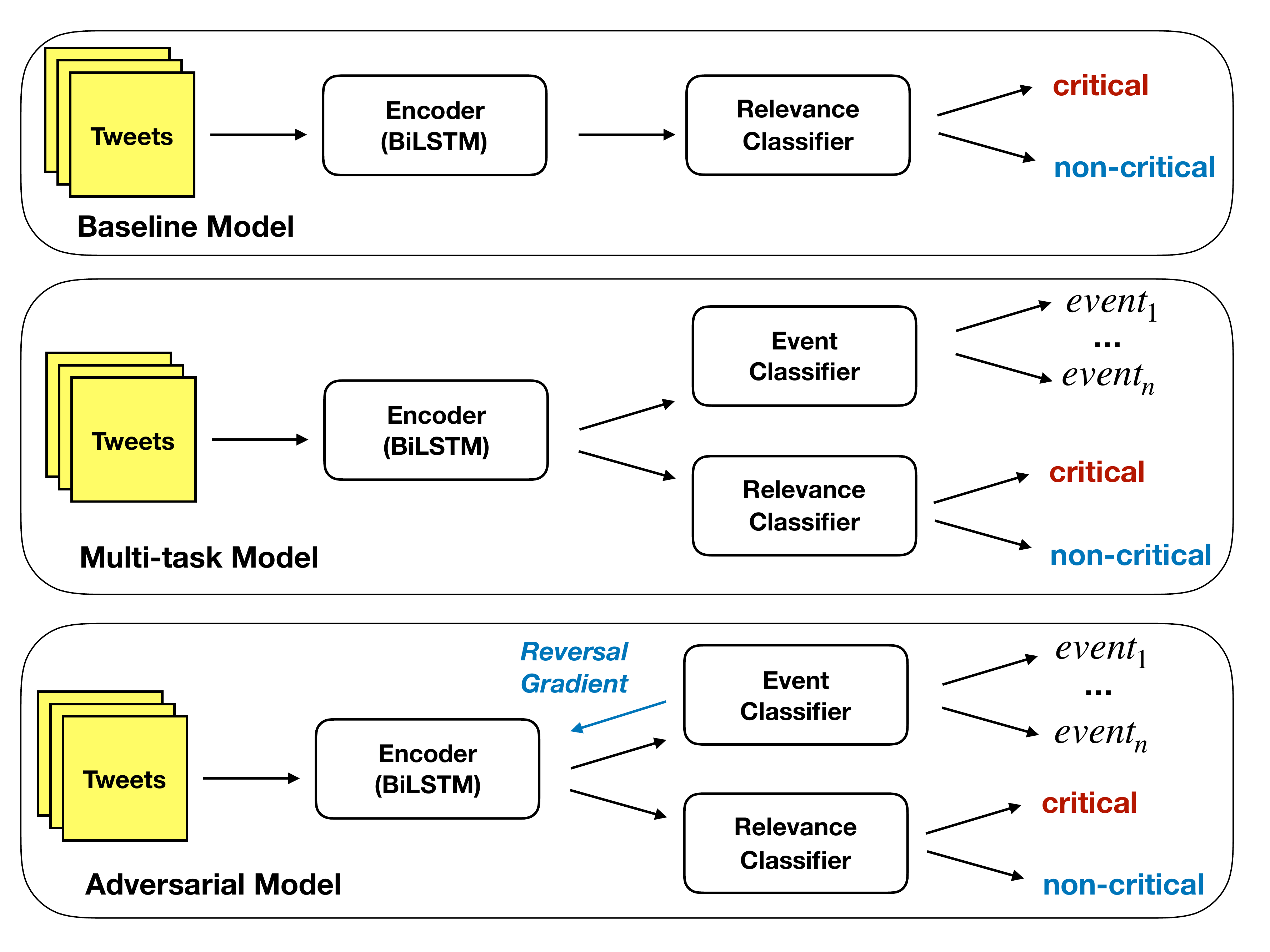}
    \caption{Evaluated Model Architectures. The adversarial model was compared against the baseline and multitask models to show the removal of event specific biases.}
    \label{fig:model_architecture}
\end{figure}

Our experimental setup consists of a dataset $\mathcal{D}$ composed of tweets $t_1,...,t_n$ and two sets of labels; $y_{e_{1}},...,y_{e_{n}}$ representing the event that the tweet belongs to and $y_{r_{1}},...,y_{r_{n}}$ representing the importance of the tweet, where $y_{r_{i}} \in \{non\mbox{-}critical, critical\}$. For this task we want to find the optimal classifier $f$ for predicting labels $y_{r_i}$. In this work we compared three models to measure if an adversarial training contributes to the detection of \textit{critical} tweets on unseen events. 

Our main hypothesis states that an adversarially trained model removes event-specific information, while focusing on features that determine how important the tweet is. For our experiments we compare the adversarially trained model against a binary classifier and a multi-task model. The comparison between the multitask and the adversarial models helps us evaluate whether the explicit removal of bias-related information benefits the relevance classifier or if using a model that jointly learns both tasks suffices.

\subsubsection{Baseline Model}

In our baseline model setup, a tweet $t_i$ is a sequence of word embeddings $w_1,...,w_{m_i}$ which are encoded through an LSTM \cite{graves2013hybrid} encoder $h$. Then the generated embedding $h(t_i)$ is fed to a binary classifier $c_r$ that learns to predict if the tweet is \textit{critical} or \textit{non-critical}. The architecture of this model is shown in Figure \ref{fig:model_architecture}.

The training loss $\mathcal{L}$ used across all the models and experiments is cross-entropy. The optimization of the baseline model is described in eq. \ref{eq:baseline_optim}.

\begin{equation}
    \label{eq:baseline_optim}
    \argmin_{h,c_r}\mathcal{L}(c_r(h(t_i)), y_{ri})
\end{equation}

\subsubsection{Multi-task Model}

The multitask learning setup described by \citet{caruana1997multitask} aims to improve the performance of a model by learning multiple tasks at the same time. Since the dataset is divided per disaster event, we take advantage of this information given by the structure of the dataset, and define event detection as the second learning task along with the criticality classification. Hence, the multitask model adds an event classifier $c_e$ on the encoding of the incoming tweet $h(t_i)$ which trains simultaneously with the classifier $c_r$, as seen in Figure \ref{fig:model_architecture}.

The optimization procedure for this model is described in eq. \ref{eq:multitask_optim}.

\begin{equation}
    \label{eq:multitask_optim}
    \argmin_{h,c_r, c_e}\mathcal{L}(c_r(h(t_i)), y_{ri}) + \mathcal{L}(c_e(h(t_i)), y_{ei})
\end{equation}

\subsubsection{Adversarial Model}

The adversarial model used in this work follows the adversarial training setup proposed by \citet{goodfellow2014generative}, \citet{ganin2016domain}, and \citet{xie2017controllable}. In essence, the adversarial model is similar to the multitask model except for the addition of a gradient-reversal layer $g_\lambda$ \cite{ganin2016domain} between the encoder $h$ and the event classifier $c_e$. The gradient-reversal layer during a forward step works as the identity function $\mathcal{I}$, but during the back-propagation step the gradient from $c_e$ is reversed and scaled by a value $\lambda$. In our work, we intend to achieve domain adaptation from previous events to a new incoming event by minimizing the information related to previously seen events provided by $c_e$, while maximizing the information gain obtained from classifier $c_r$, as described in eq. \ref{eq:adversarial_optim}.

\begin{equation}
    \label{eq:adversarial_optim}
    \argmin_{h,c_r, c_e}\mathcal{L}(c_r(h(t_i)), y_{ri}) + \mathcal{L}(c_e(g_\lambda(h(t_i))), y_{ei})
\end{equation}

\section{Experiments}

\begin{table*}[ht]
\centering
\caption{Event based zero-shot test results. The best model per disaster type is highlighted with the color assigned to the disaster type. The best model per embedding type is highlighted in bold.}
\resizebox{0.8\textwidth}{!}{
    \begin{tabular}{lllccc}
    \toprule
    \textbf{Event Type} & 
    \textbf{Embedding} & 
    \multicolumn{1}{l}{\textbf{Model}} & \textbf{\begin{tabular}[c]{@{}c@{}}Macro\\ F1\end{tabular}} & \textbf{\begin{tabular}[c]{@{}c@{}}Non-Critical\\ F1\end{tabular}} & \textbf{\begin{tabular}[c]{@{}c@{}}Critical\\ F1\end{tabular}} \\ \midrule
    \multirow{6}{*}{Earthquakes} & \multirow{3}{*}{GloVe} 
        & Baseline & 0.6432 & 0.9082 & 0.3782 \\ 
     &  & Multitask & 0.5890 & 0.8960 & 0.2819 \\ 
     &  & \cellcolor{earthquake}Adversarial & \cellcolor{earthquake}\textbf{0.6602} & \cellcolor{earthquake}0.9170 & \cellcolor{earthquake}\textbf{0.4034} \\ \cmidrule{2-6}
     & \multirow{3}{*}{BERT} 
        & Baseline & 0.6138 & 0.9062 & 0.3213 \\ 
     &  & Multitask & 0.5844 & 0.8863 & 0.2826 \\ 
     &  & Adversarial & \textbf{0.6154} & 0.8888 & \textbf{0.3420} \\ \midrule
     
    \multirow{6}{*}{Floods} & \multirow{3}{*}{GloVe} 
        & Baseline & 0.6010 & 0.8674 & 0.3346 \\ 
     &  & Multitask & 0.6130 & 0.8679 & 0.3581 \\ 
     &  & \cellcolor{flood}Adversarial & \cellcolor{flood}\textbf{0.6326} & \cellcolor{flood}0.8454 & \cellcolor{flood}\textbf{0.4198} \\ \cmidrule{2-6}
     & \multirow{3}{*}{BERT} 
        & Baseline & 0.6145 & 0.8834 & 0.3455 \\ 
     &  & Mulitask & 0.6062 & 0.8793 & 0.3331 \\ 
     &  & Adversarial & \textbf{0.6403} & 0.8642 &\textbf{0.4164} \\ \midrule
     
     \multirow{6}{*}{Typhoons} & \multirow{3}{*}{GloVe} 
        & Baseline& 0.5714 & 0.8965 & 0.2462 \\ 
     &  & Multitask& 0.5832 & 0.8961 & 0.2702 \\ 
     &  & Adversarial& \textbf{0.5887} & 0.8916 & \textbf{0.2858} \\ \cmidrule{2-6}
     & \multirow{3}{*}{BERT} 
        & Baseline & 0.6249 & 0.9189 & 0.3310 \\ 
     &  & Mulitask & 0.6291 & 0.9091 & 0.3491 \\ 
     &  & \cellcolor{typhoon}Adversarial & \cellcolor{typhoon}\textbf{0.6302} & \cellcolor{typhoon}0.9086 & \cellcolor{typhoon}\textbf{0.3517} \\ \midrule

    \multirow{6}{*}{Attacks} & \multirow{3}{*}{GloVe} 
        & Baseline & 0.6049 & 0.9047 & 0.3052 \\ 
     &  & Multitask & 0.5994 & 0.8917 & 0.3071 \\ 
     &  & Adversarial & \textbf{0.6056} & 0.8975 &\textbf{0.3137} \\ \cline{2-6}
     & \multirow{3}{*}{BERT} 
        & Baseline  & 0.5744 & 0.8840 & 0.2649 \\ 
     &  & \cellcolor{attack}Multitask & \cellcolor{attack}\textbf{0.6165}  & \cellcolor{attack}0.9009 & \cellcolor{attack}\textbf{0.3322} \\ 
     &  & Adversarial& 0.5492 & 0.8511 & 0.2472 \\ 
     \bottomrule
    \end{tabular}
}
\label{tab:comparison_results}
\end{table*}

For our experiments we used two of the main popular word embeddings to represent the tokens of the tweets in the target dataset: GloVe \cite{pennington2014glove} embeddings, and BERT \cite{devlin2019bert} embeddings.

We used the 100-dimensional GloVe embeddings pre-trained on Wikipedia and Gigaword, which were made publicly available by the authors$^3$\footnote{$^3$ \href{https://nlp.stanford.edu/projects/glove/}{https://nlp.stanford.edu/projects/glove/}}. For extracting BERT embeddings we used the Python package \textit{bert-embeddings}$^4$\footnote{$^4$ \href{https://github.com/imgarylai/bert-embedding}{https://github.com/imgarylai/bert-embedding}} as we built the networks for our experiments in PyTorch. This package offers a pre-trained 768-dimensional hidden state transformer model with 12-layers and 12-headed attention. In our experiments, the BERT model was frozen with no fine-tuning during training.

Throughout all of our experiments the tweet encoder $h$ is an LSTM with two layers. Each of the LSTMs have a hidden dimension of 100, which results in a tweet embedding of size 200. Both classifiers $c_r$ and $c_e$ are linear layers with output size 2 and the number of events per experiment, respectively. During our initial experimentation, we set the gradient-reversal layer scaling value lambda to different values within the range $[0.1-10]$. The most stable result throughout the whole experiments was obtained with $\lambda=1$.

The models were trained using the Adam optimizer \cite{kingma2014adam}, with an initial learning rate 0.01, batch size 16 and trained for 40 epochs. We employed dynamic batching by padding each batch to the sequence length of the longest sample in the batch. 

To test the performance of the model at every epoch we calculated the micro F1 on the \textit{critical} class from $c_r$ and considered as the best model the one which showed the highest Critical-F1 score, since for disasters it is important to recall as many \textit{critical} tweets with the highest possible precision.

\subsection{Model Evaluation}

Since we intend to evaluate the models for a real-life scenario, we used data from each disaster type separately (e.g. model trained and tested only on flood events), to perform an analysis in a disaster-based zero-shot learning scenario simulating an incoming unseen event. To achieve this, the training data consists of all the events of the same disaster type except one, as it is used for testing the model. We generated $n$ splits for each event type, where $n$ is the amount of events per event type. We evaluated the three models on each split obtaining the macro-F1 and the micro-F1 scores from the $c_r$ predictions. Finally, we calculated the mean of these metrics, which we can see in Table \ref{tab:comparison_results}. The best models for each event type are highlighted in the representative color of the event, as shown in Figure \ref{fig:dataset_pie}.

Since we follow a leave-one-out testing procedure, we could not include the wildfires event type since this category only has two instances. This makes it impossible to train the multitask and adversarial models on this type of event.

Our experiments show an improvement of the F1 score for all disaster events that use adversarial training except for the attacks group, where the improvement is not consistent with the rest of the events. The earthquake and flood events show a significantly better performance of the adversarial model when compared to both the baseline and the multitask model. For the typhoon events the multitask model improves slightly over the baseline, but the adversarial model is the best for both embedding types, while BERT has better results than GloVe by a large margin.

Most similar to our setting, \citet{nguyen2016applications} performs an experiment in an online training scenario using the Nepal 2015 Earthquake as test set, while more than 10,000 tweets from the dataset are used for pre-training the model. Their work reports an AUC of 0.73 at the beginning of the event, which would be comparable to our zero-shot learning scenario. To compare our model to their work, we used the data split where the Nepal earthquake was left out for testing the model. On this data split, the adversarial model using BERT embeddings obtains an AUC of 0.62 for the critical class while training with only 815 tweets from all the other earthquake events.

\subsection{Event Types Data Mix}
\begin{table}[]
\caption{Mixed flood and typhoon test results}
\resizebox{\columnwidth}{!}{
    \begin{tabular}{lccc}
    \toprule
    \multicolumn{1}{l}{\textbf{Model}} & \textbf{Macro F1} & \textbf{Non-Critical F1} & \textbf{Critical F1} \\ \midrule
    Baseline - GloVe & \textbf{0.5376} & 0.7602 & 0.\textbf{3150} \\
    MultiTask - GloVe & 0.5331 & 0.7529 & 0.3133 \\
    Adversarial - GloVe & 0.5157 & 0.7428 & 0.2885 \\ \midrule
    Baseline - BERT & 0.5593 & 0.7602 & 0.3584 \\
    MultiTask - BERT & \textbf{0.5625} & 0.7558 & 0.\textbf{3692} \\
    Adversarial - BERT & 0.5539 & 0.7500 & 0.3578 \\ 
    \bottomrule
    \end{tabular}
}
\label{tab:water_results}
\end{table}

In Figure \ref{fig:dataset_pie}, we observe that the attack events group consists of diverse types of events such as shootings, bombings, and explosions. Even though all of those events contain violence-related incidents, the adversarial model with BERT embeddings has lower performance than the baseline and the multitask learning model, as shown in the results on Table \ref{tab:comparison_results}. Our hypothesis is that the adversarial model fails to remove the event-specific biases in the Attack group, because of the mixture of different event types. A potential solution to this problem would be to include more events to facilitate the disentanglement of the Attacks group.

To test this hypothesis, we created a synthetic event type where we mix flood and typhoon events, since both are disasters that would result in flooded cities and towns. We repeated the same experimental procedure by leaving out one event for testing and obtained the mean scores across all splits, as reported in Table \ref{tab:water_results}. The results from this experiment verify our hypothesis that the adversarial training of the classifier is sensitive to the entanglement of events in the training data. This supports our claim on why we have low performance on attacks and highlights the importance of not mixing different event types when training under an adversarial setup.

\section{Qualitative Analysis}

\begin{table*}[]
\centering
\caption{Examples captured by the adversarial model (true-positives), but not the baseline (false-negatives).}
\begin{tabular}{ll}
\toprule
\multicolumn{1}{l}{\textbf{True Label}} & \multicolumn{1}{l}{\textbf{Tweet Text}} \\ \midrule
\rowcolor[HTML]{EFEFEF}
Critical & \begin{tabular}[c]{@{}l@{}}rt flood in the ust hospital is now on the 2nd floor\\ no food for the patients \& staff pls help ...\end{tabular} \\
Critical & rt please help rt rt those who are in u erm the flood is now goi ... \\
\rowcolor[HTML]{EFEFEF}
Critical & \begin{tabular}[c]{@{}l@{}}ust hospital and u erm in need of immediate help u sts\\ morgue is flooded ue rms nursery is near being flooded please please\end{tabular} \\
Critical & philippine flood fatalities hit 23 \\ \midrule
\rowcolor[HTML]{EFEFEF}
Non-Critical & \begin{tabular}[c]{@{}l@{}}metro manila flood updates nlex is now north luzon express river\\ pls rt and spread\end{tabular} \\
Non-Critical & \begin{tabular}[c]{@{}l@{}}ndr rmc nearly 50 of metro manila submerged in floodwater\\ due to heavy monsoon rains\end{tabular} \\
\rowcolor[HTML]{EFEFEF}
Non-Critical & \begin{tabular}[c]{@{}l@{}}rt lets all pray for those who lost their homes and now living in\\ cold and starving ...\end{tabular} \\
Non-Critical & \begin{tabular}[c]{@{}l@{}}rt pal passengers to/from manila who are unable to take\\ their flights due to floods may rebook their tickets with rebooking c ...\end{tabular} \\ 
\bottomrule
\end{tabular}
\label{tab:qualitative_TP}
\end{table*}

We took a deeper look into our experimental results by comparing which patterns are learnt by the adversarial model but not the baseline. For this analysis, we focused on flood and earthquake event types, as they show the greatest difference in F1 score between the baseline and the adversarial model.

\begin{table}[!]
\caption{Test results on 2012 Philippines Flood}
\resizebox{\columnwidth}{!}{
    \begin{tabular}{lccc}
    \toprule
    \multicolumn{1}{l}{\textbf{Model}} & \textbf{Macro F1} & \textbf{Non-Critical F1} & \textbf{Critical F1} \\ \midrule
    Baseline - BERT & 0.5844 & 0.8413 & 0.3274 \\
    MultiTask - BERT & 0.5875 & 0.8766 & 0.2985 \\
    Adversarial - BERT & \textbf{0.6535} & 0.8832 & \textbf{0.4238} \\ \bottomrule
    \end{tabular}
}
\label{tab:flood_philippines_results}
\end{table}

\subsection{Critical Detection Comparison}


For the first part of the qualitative analysis, we examined tweets where the baseline and the adversarial models disagree upon. We looked at both critical and non-critical tweets in order to find common patterns where the models fail. In Table \ref{tab:qualitative_TP} we show some examples of tweets where the baseline model failed, but were correctly classified by the adversarial model. The examples used come from the Philippines flood (performance shown in Table \ref{tab:flood_philippines_results}). 

A consistent pattern observed for the critical tweets is that they mostly contain information about a need for emergent help or a situation currently happening. Furthermore, we see a strong sentiment of despair, where we may assume that the users are directly affected by the event. On the other hand, if we look at the non-critical tweets that were incorrectly classified as critical by the baseline, they mostly contain location information and named entities. As mentioned earlier, in a zero-shot scenario upon the development of a crisis event, the models trained on previous similar scenarios perform poorly due to event bias found in the data. Through those examples we see that our approach successfully removes part of that bias through adversarial learning.

\subsection{Model Comparison via Saliency Maps}

\begin{figure*}[hbt!]
    \centering
    \includegraphics[trim=0 80 0 70, clip, width=0.9\textwidth]{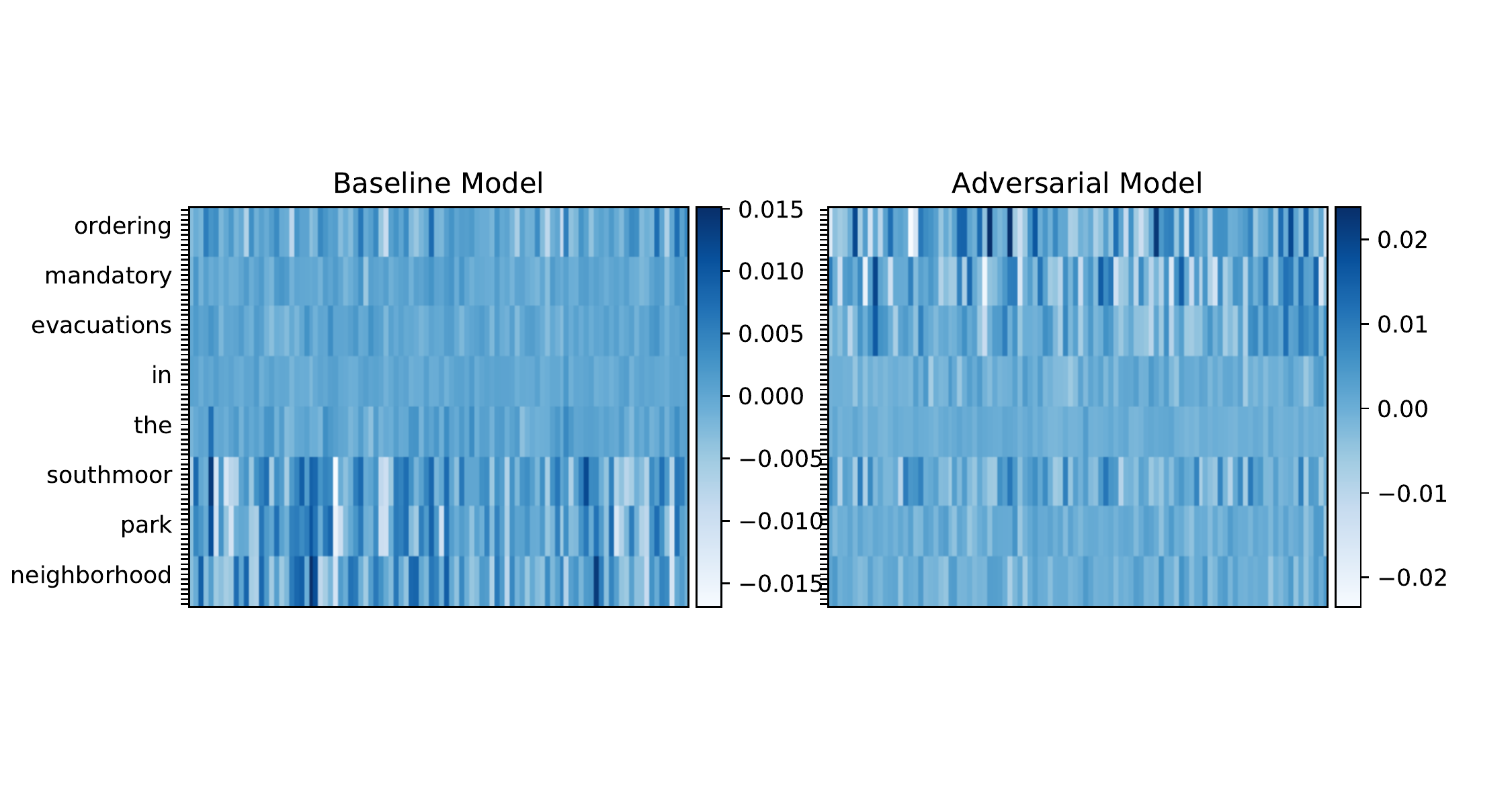}
    \includegraphics[trim=0 70 0 60, clip, width=0.9\textwidth]{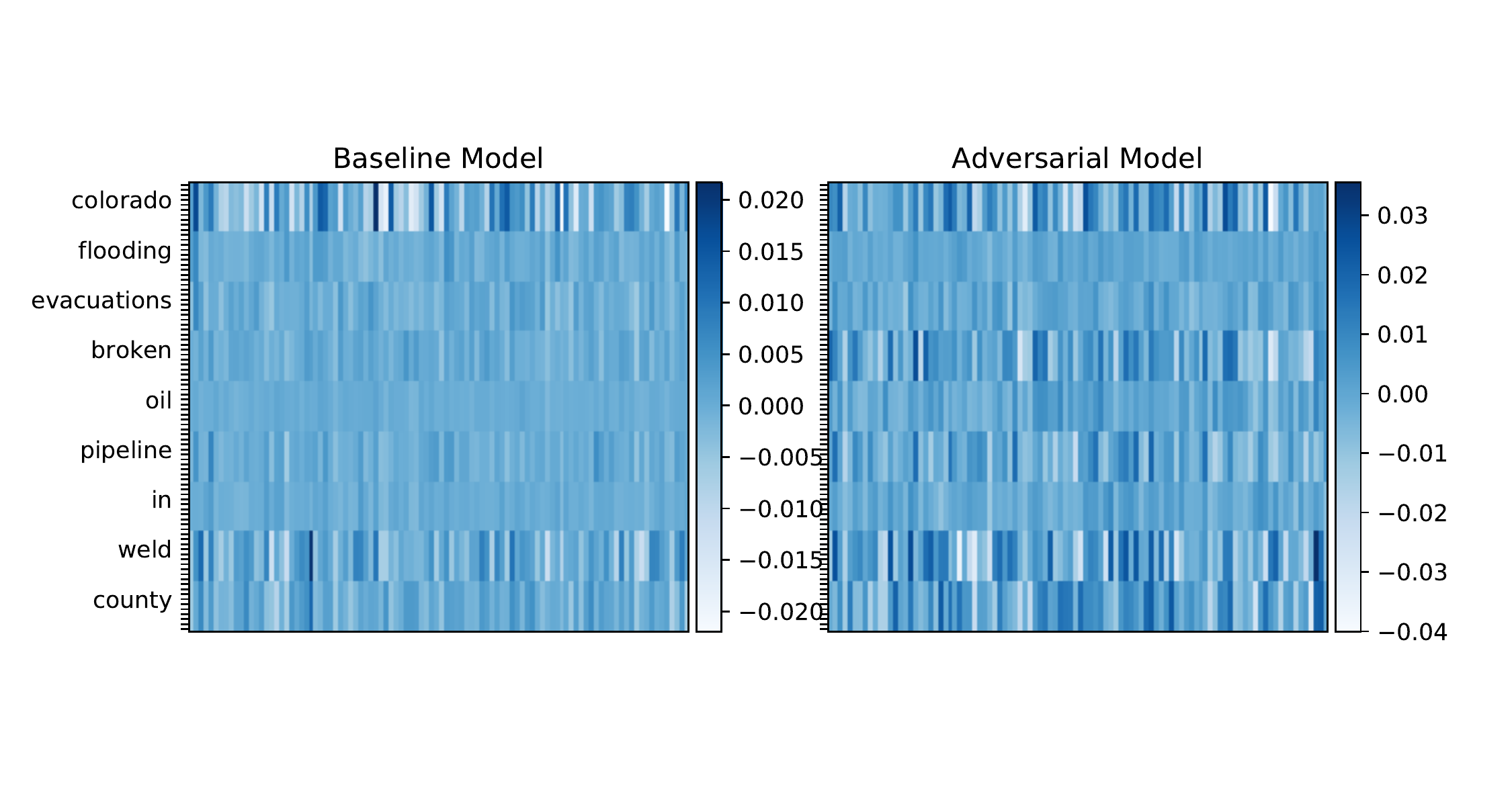}
    \includegraphics[trim=0 50 0 35, clip, width=0.9\textwidth]{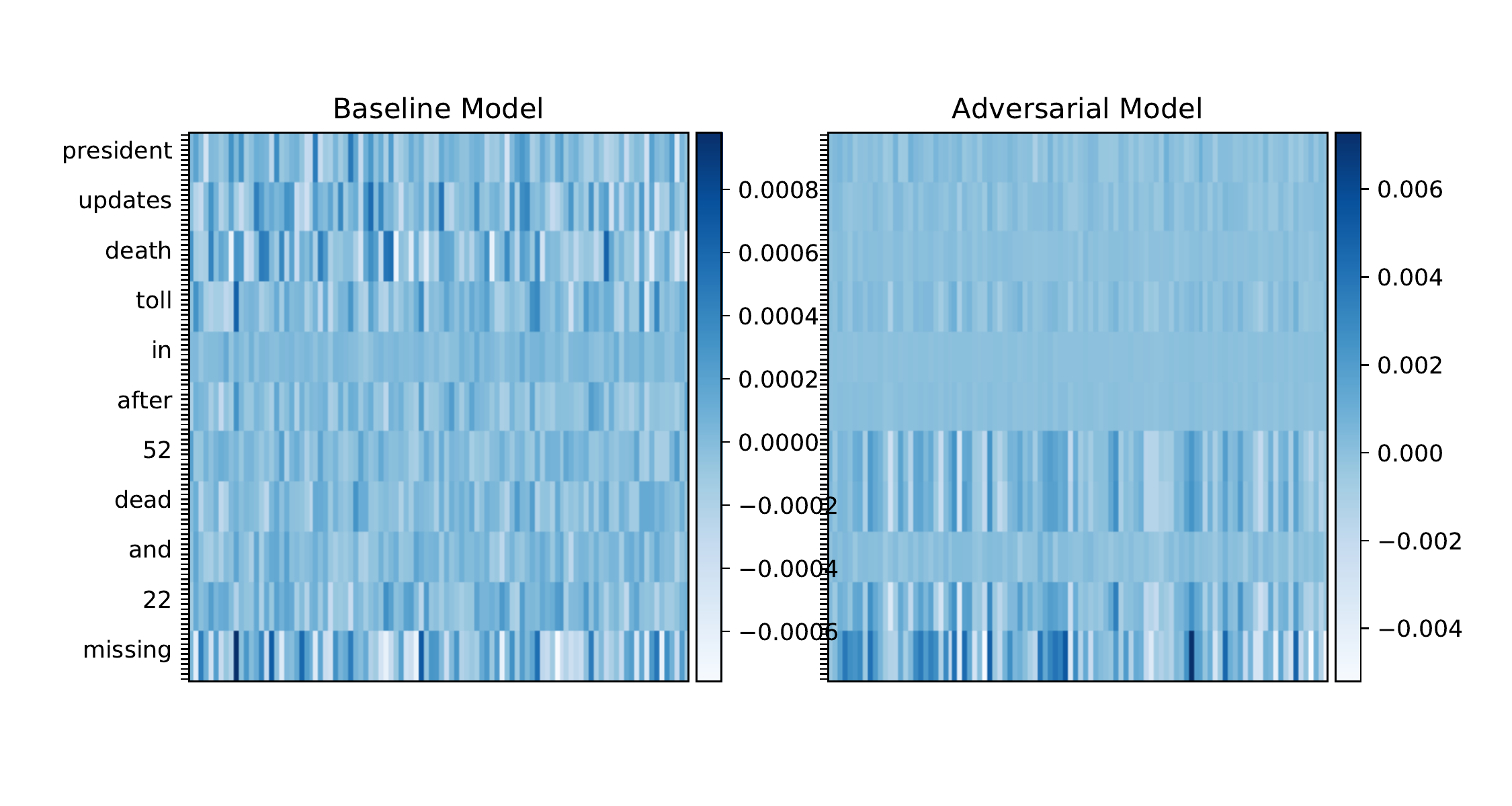}
    \caption{Saliency map visualization of tweets with strong event-bias.}
    \label{fig:vis_embeds}
\end{figure*}
For the second part of our analysis, we used saliency maps to visualize the relevance of each word in a tweet for the models. We selected tweets that contain named entities (e.g. locations, names) or information that is generally important to classify a tweet, such as casualties. For this part, we only used GloVe embeddings, since BERT is context-based and each embedding may encode information from the rest of the tweet.

In order to construct the saliency map, we used back-propagation to estimate the first-order derivatives from each word, as a measure of their contribution to the model's decision. This strategy was adopted from the vision community \cite{erhan2009visualizing,simonyan2013deep}, and recently adapted in NLP research \cite{li2016visualizing}.

In Figure \ref{fig:vis_embeds} we visualize the saliency map of each word embedding for the baseline and adversarial models. The higher the absolute value of the first-order derivative (dark blue and white), the more important role it plays into the classifier's decision. We observe that, for the first and second sentences, the baseline puts more weight on the location, which is a strong event-bias since it includes information only for a particular event and not a disaster type (e.g. floods). On the other hand, the adversarial model focuses more on important sub-events, like \textit{mandatory evacuations} and \textit{broken pipeline}, which we desire to capture in a zero-shot scenario, and is generally ignored by the baseline model. We further observe a similar trend for the third sentence, where the baseline gives mostly uniform weight with a small focus on \textit{president updates death}, while the adversarial model focuses more on generally informative text that describes casualties.

\section{Future Work}

Our experiments show that mixing data from events whose semantics are similar, like the violent mass attacks and the synthetically generated set of floods and typhoons, confuses the adversarial model. As a result, it does not show any improvement over the baseline. Moreover, in some cases models trained with GloVe achieved better performance compared to those trained with BERT. For this reason, it seems appropriate to fine-tune transformer-based language models so we could take advantage of the large amount of unlabeled data provided by the Crisis NLP dataset that was not used in this work. 

Given that our ultimate goal is to detect and use actionable information during crisis events to inform life-saving actions, an essential part of future research is to design interpretable models. An interesting work proposes a new approach to interpretable classification named deep weighted averaging classifiers (DWAC) \cite{card.2019.dwac}, which gives an explanation of the prediction in terms of the weighted sum of training instances. DWAC could replace the importance classifier $c_r$ in our proposed adversarial model. An advantage of using DWAC is that it would deliver the most relevant tweets from the training data which contributed to the detection of a critical tweet.


Finally, since we deal with a real-time information stream it seems appropriate to evaluate this model in an online learning scenario \cite{nguyen2016applications}.

\section{Conclusion}

In this work, we compared an adversarialy trained model against a baseline classifier and a multitask learning model. The main task for all the models was to predict if a tweet is \textit{critical} or \textit{non-critical} over four types of disaster events: earthquakes, floods, typhoons, and mass attacks in public spaces. We presented a thorough analysis on how a simple classification model trained on crisis event data can be improved through adversarial training. Our results showed how the addition of an adversarial network removes the bias from specific events, allowing the network to put more attention in disaster related information rather than specificities of a particular event. In most of our experiments the adversarially trained model obtained the highest F1 score.

Our experimental results demonstrate the relevance of using micro-F1 scores for evaluating the detection of critical posts from an information stream such as Twitter. The impact of false negatives while detecting critical tweets is larger than the false positives, since we would be missing decisive information from the data stream. Hence, micro-F1 score is a more informative metric to consider instead of accuracy, or even the overall F1 score since event crisis detection usually suffers from highly skewed data towards the irrelevant samples of the dataset.

\section*{Acknowledgments}

This research was partially supported by DARPA grant no HR001117S0017-World-Mod-FP-036 funded under the World Modelers program, as well as by the financial assistance award 60NANB17D156 from U.S. Department of Commerce, National Institute of Standards and Technology (NIST). The U.S. Government is authorized to reproduce and distribute reprints for Governmental purposes notwithstanding any copyright annotation/herein. Disclaimer: The views and conclusions contained herein are those of the authors and should not be interpreted as necessarily representing the official policies or endorsements, either expressed or implied, of NIST, DOI/IBC, or the U.S. Government.

\bibliography{anthology,emnlp2020}
\bibliographystyle{acl_natbib}

\end{document}